\documentclass[letterpaper]{article} 
\usepackage{aaai25}  
\usepackage{times}  
\usepackage{helvet}  
\usepackage{courier}  
\usepackage[hyphens]{url}  
\usepackage{graphicx} 
\usepackage{natbib}  
\usepackage{caption} 
\usepackage{algorithm}
\usepackage{algorithmic}
\usepackage{amsmath}
\usepackage{xcolor}
\usepackage{newfloat}
\usepackage{listings}
\DeclareCaptionStyle{ruled}{labelfont=normalfont,labelsep=colon,strut=off} 
\floatstyle{ruled}
\newfloat{listing}{tb}{lst}{}
\floatname{listing}{Listing}
\title{FlowPolicy: Enabling Fast and Robust 3D Flow-based Policy via Consistency Flow Matching for Robot Manipulation}
\author{
    Qinglun Zhang\textsuperscript{\rm 1,2,\equalcontrib},
    Zhen Liu\textsuperscript{\rm 1,2,\equalcontrib},
    Haoqiang Fan\textsuperscript{\rm 2},
    Guanghui Liu\textsuperscript{\rm 1},
    Bing Zeng\textsuperscript{\rm 1},
    Shuaicheng Liu\textsuperscript{\rm 1,2,\footnote{Corresponding authors}}
}
\affiliations{
    \textsuperscript{\rm 1}University of Electronic Science and Technology of China \\
    \textsuperscript{\rm 2}Megvii Technology \\
    \{zhangqinglun26@std., liuzhen03@std., guanghuiliu@, eezeng@, liushuaicheng@\}uestc.edu.cn, fhq@megvii.com
}

\usepackage{bibentry}

\usepackage{booktabs}
\usepackage{multirow}
\usepackage{multicol}
\usepackage{enumerate}
\usepackage{array}

\usepackage{amsmath,amsfonts}
\usepackage{algorithmic}
\usepackage{algorithm}
\usepackage{array}
\usepackage{textcomp}
\usepackage{url}
\usepackage{verbatim}
\usepackage{graphicx}
\usepackage{cite}
\usepackage{bbding}
\usepackage{color}

\begin{document}

\maketitle

\begin{abstract}

Robots can acquire complex manipulation skills by learning policies from expert demonstrations, which is often known as vision-based imitation learning. Generating policies based on diffusion and flow matching models has been shown to be effective, particularly in robotic manipulation tasks. However, recursion-based approaches are inference inefficient in working from noise distributions to policy distributions, posing a challenging trade-off between efficiency and quality. This motivates us to propose \textbf{FlowPolicy}, a novel framework for fast policy generation based on consistency flow matching and 3D vision. Our approach refines the flow dynamics by normalizing the self-consistency of the velocity field, enabling the model to derive task execution policies in a single inference step. Specifically, FlowPolicy conditions on the observed 3D point cloud, where consistency flow matching directly defines straight-line flows from different time states to the same action space, while simultaneously constraining their velocity values, that is, we approximate the trajectories from noise to robot actions by normalizing the self-consistency of the velocity field within the action space, thus improving the inference efficiency. We validate the effectiveness of FlowPolicy in Adroit and Metaworld, demonstrating a 7$\times$ increase in inference speed while maintaining competitive average success rates compared to state-of-the-art methods. Code is available at https://github.com/zql-kk/FlowPolicy.

\end{abstract}

%

\section{Introduction}

Enabling robots to acquire diverse manipulation tasks by learning from expert demonstrations, commonly referred to as imitation learning, has been a long-standing objective in robotic learning and embodied AI~\cite{atkeson1997robot,argall2009survey,wang2024diffail,zare2024survey,li2024robust}. Prior work attempts to tackle this problem either by explicitly defining it as a supervised regression task~\cite{zhang2018deep,florence2019self,toyer2020magical,rahmatizadeh2018vision}, or by implicitly modeling action distributions using energy-based models~\cite{florence2022implicit,jarrett2020strictly}. The former learns the mapping from observations to actions, while the latter seeks to identify the minimal energy action through energy optimization. However, explicit regression approaches are often inadequate for capturing multi-modal demonstration behaviors and struggle with high-precision tasks~\cite{florence2022implicit}. Conversely, energy-based models face challenges with training stability, primarily due to the necessity of negative sample extraction during the training process~\cite{chi2023diffusion}.

\begin{figure}[t]
\centering
\includegraphics[width=1.0\linewidth]{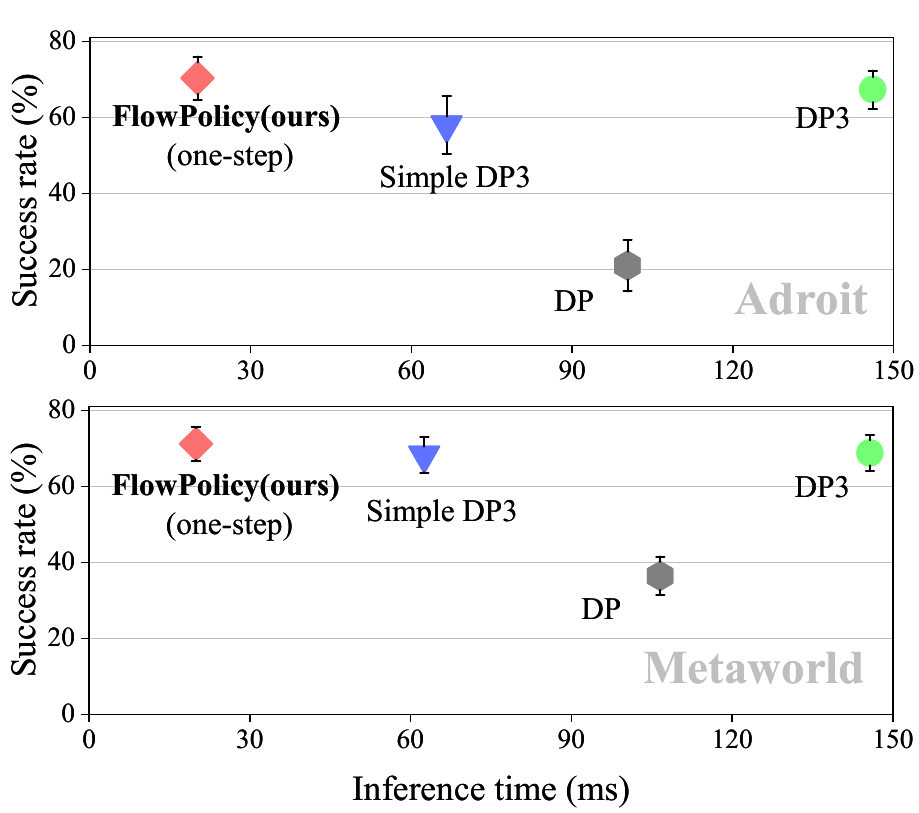}
\caption{Comparison of FlowPolicy with the state-of-the-art 2D-based method DP\cite{chi2023diffusion} and 3D-based methods DP3~\cite{ze20243ddiffusionpolicygeneralizable} and its lightweight version Simple DP3 in terms of inference time and average success rate on Adroit and Metaworld.}
\label{fig:teaser}
\end{figure}

Recent advancements in diffusion-based policy learning frameworks~\cite{wang2022diffusion,pearce2023imitating,reuss2023goal,chi2023diffusion,ze20243ddiffusionpolicygeneralizable} have elevated this task to new heights, leveraging their powerful capability to model complex, high-dimensional robotic trajectories. Within this context, Diffusion Policy~\cite{chi2023diffusion} defines policy learning as a conditional denoising diffusion process over the robot action space, conditioned on 2D observation features. The subsequent 3D Diffusion Policy~\cite{ze20243ddiffusionpolicygeneralizable}, also known as DP3, further extends the conditional input to more complex 3D visual representations and is thus capable of acquiring generalizable visuomotor policies with only a few dozen expert demonstrations. Nevertheless, diffusion-based solutions are inevitably plagued by substantial runtime inefficiencies, as they typically require numerous sampling steps during inference to generate high-quality actions. While various sampling acceleration techniques~\cite{song2020denoising,song2023consistency} have been proposed, the critical challenge of balancing efficiency and policy quality persists, severely limiting the practical application of these learned policies.

On the other hand, Flow Matching (FM)~\cite{lipman2022flow}, another class of generative model frameworks that directly defines probability paths via Ordinary Differential Equations (ODEs) to transform between noise and data samples, has shown greater numerical stability and inference efficiency than diffusion families. The latest advancement in this area, Consistency Flow Matching (Consistency-FM)~\cite{yang2024consistencyflowmatchingdefining}, further improves efficiency by defining a straight-line flow from any point in time to the same endpoint, making one-step generation possible, which offers a potential direction for real-time policy generation. However, the conditional generation capability of Consistency-FM, particularly for embodied tasks involving complex 3D vision representations as conditions, remains underexplored.

In this paper, we address these challenges in policy generation by leveraging the concept of consistency flow matching, introducing a novel 3D flow-based framework for real-time robotic manipulation policy generation, named FlowPolicy. More specifically, we make the first attempt at conditional consistent flow matching in 3D robotic manipulation tasks. FlowPolicy allows for the fastest possible generation of target actions by directly defining straight-line flow conditions on the initial 3D point clouds. In addition, multi-segment training is applied to ensure the quality of action generation, aiming at a better trade-off between efficiency and effectiveness. We evaluated our framework on 37 robot manipulation tasks from Adroit and Metaworld. 
As illustrated in Fig.~\ref{fig:teaser}, our approach achieves a 7$\times$ reduction in average inference time while maintaining a competitive average success rate compared to state-of-the-art methods based on 3D visual conditional representation. By combining consistency flow matching with 3D condition-based robot manipulation, we enable the real-time decoding of high-quality robot actions with one-step inference.
\begin{figure*}[ht]
\centering
\includegraphics[width=1\textwidth]{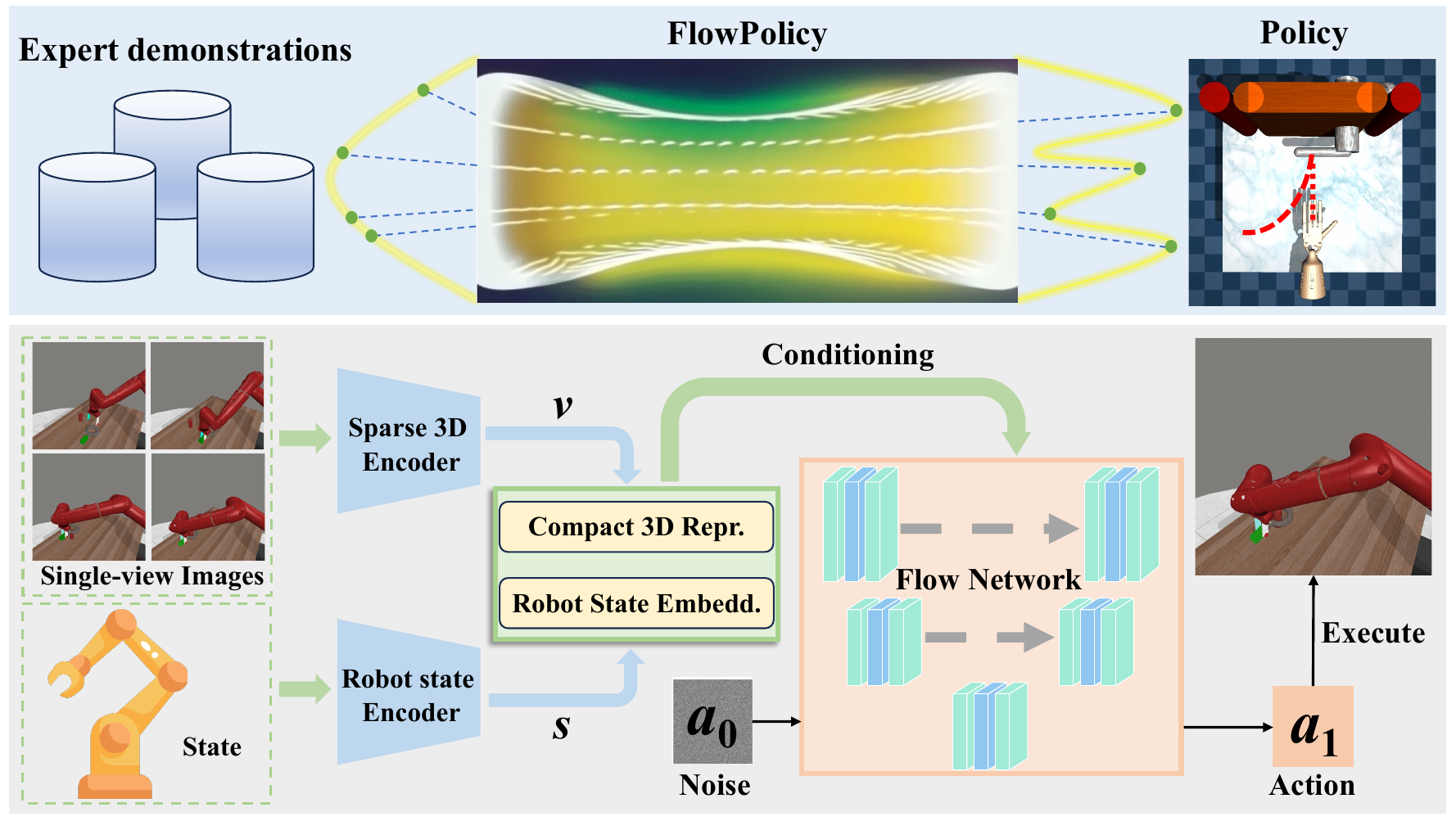}\\
\caption{\textbf{Overall pipeline}. The top section visualizes FlowPolicy, where a straight-line flow enables the fastest data transition from the noise distribution to the action distribution (Adroit: Open the door). The bottom section shows the details of FlowPolicy: expert demonstrations are converted to 3D point clouds, which, along with the robot state, are encoded into compact 3D visual representations and state embeddings. A straight-line flow is then learned via conditional consistency flow matching, generating high-quality actions for tasks (Metaworld: Assembly) at real-time inference speed.}

\label{fig:pipeline}
\end{figure*}
In summary, our main contributions are threefold:
\begin{itemize}
    \item We first propose a 3D flow-based policy generation framework that conditions the 3D visual representation and can generate robust robotic actions with few demonstrations, namely \textbf{FlowPolicy}.
    \item The proposed FlowPolicy significantly narrows the gap between efficiency and quality in vision-based imitation learning algorithms, enabling \textbf{real-time policy generation} across diverse manipulation tasks.
    \item We conduct extensive experiments on 37 simulation tasks from Adroit and Metaworld, demonstrating that our approach achieves \textbf{state-of-the-art average success rates}.
\end{itemize}

\section{Related Work}

\subsection{Application of Diffusion Models to 3D Robotics } 
Diffusion models are a family of generative models that are highly generalizable and easy to train, which are initially utilized for image generation tasks~\cite{batzolis2021conditional,epstein2023diffusion}. As a subversive approach, diffusion models progressively refine Gaussian noise by recursively refining it to the point where it satisfies the data distribution. Recently, it has been introduced to the field of robotics, including reinforcement learning~\cite{romer2024safe}, imitation learning~\cite{wang2024rise}, trajectory optimization~\cite{carvalho2023motion} and grasping~\cite{urain2023se}, and have shown impressive performance. Denoising Diffusion Probabilistic Models (DDPMs)~\cite{ho2020denoising} generate a clean sample from noise by solving a Stochastic Differential Equation (SDE). However, numerical simulations of diffusion processes typically involve hundreds of steps, greatly slowing the inference speed. Denoising Diffusion Implicit Models (DDIMs)~\cite{song2020denoising} speeds up inference by solving the inverse process as an Ordinary Differential Equation (ODE), but this sacrifices generation quality. Although adaptive ODE solvers are available, they do not meaningfully reduce the number of inference steps. In 3D robot manipulation tasks involving high-dimensional control, achieving real-time inference speed with high-quality actions using diffusion models is challenging. The primary obstacles arise from the inherent uncertainty of the diffusion process and its recursive nature. 

To improve the efficiency of the diffusion process, Song~\cite{song2023consistency} proposes the consistency model. The consistency model directly maps noise to a data distribution by learning to map points at arbitrary moments on the probabilistic flow ODE (PF-ODE) trajectory to the origin (i.e., clean data) to support one-step or several-step generation. It has been utilized in several domains such as high-fidelity image generation~\cite{luo2023latent}, text-to-3D generation~\cite{seo2023let}, and video editing~\cite{chai2023stablevideo}. Although consistency models support one-step generation~\cite{lu2024manicm}, they typically require distillation to achieve the desired results, which restricts their applicability in real-world robotics. Compared to the above methods, our FlowPolicy does not require numerical simulation or distillation.

\subsection{Flow Matching}
Flow Matching (FM)~\cite{lipman2022flow} is an emerging family of generative models based on optimal transport theory~\cite{mozaffari2016optimal}, which aims to estimate a vector field that yields an ODE and admit the desired probabilistic path. FM allows direct regression to the goal of estimating the vector field, and hence it is simpler and more numerically stable than DDPM and DDIM. By avoiding estimating noise and instead matching a path from the noise to the target, FM enables faster inference, which is crucial in real-time robot manipulations. Previous works have demonstrated the effectiveness of FM for inverse reinforcement learning~\cite{chang2022flow}, multi-support motion~\cite{rouxel2024flow}, and robot deployment~\cite{feng2023topologymatchingnormalizingflowsoutofdistribution}. However, FM-based generation of complex robot action has rarely been reported, especially involving complex 3D spatial representations. In this work, we explore the feasibility of applying FM to 3D robot manipulation and hope to achieve real-time generation through consistency constraints.

Recently, Yang~\cite{yang2024consistencyflowmatchingdefining} proposed to perform consistency in velocity field space rather than sample space to improve the flow matching process (Consistency-FM). Consistency-FM directly defines a straight-line flow from any time to the same endpoint and imposes constraints on its velocity values. Also, the multi-segment training technique allows Consistency-FM to achieve a better trade-off between sampling quality and efficiency. Therefore, Consistency-FM is an elegant balance of efficiency and performance. More importantly, 
Consistency-FM can be trained to produce a robust flow model without the aid of distillation, which is valuable to robots performing unseen tasks, as it is difficult to find a robust teacher model in practical applications. While consistency-based models have shown promising results in some areas of embodied intelligence~\cite{ding2024consistencymodelsrichefficient,chen2024boostingcontinuouscontrolconsistency,prasad2024consistencypolicyacceleratedvisuomotor}, its effectiveness has not been demonstrated on embodied tasks with complex 3D vision representation as conditions, especially for flow-based methods. To address this issue, we propose FlowPolicy, 
a real-time 3D policy generation framework based on consistency flow matching.

\section{Method}
Our method expects a limited number of expert demonstrations to teach an agent to learn a policy $\pi:O \Longrightarrow A$, i.e., mapping from visual observations $\textit{o}\in{O}$ to actions $\textit{a}\in{A}$. Visual observations include the robot state and scene point clouds, and actions are usually sequences of trajectories of the robot to accomplish a specific task. Diffusion-based and flow-based methods can generate high-quality robot actions, but expensive iterative sampling makes them unable to accomplish real-time operations, and thus the application scenarios are very limited. While Consistency-FM balances efficiency and effectiveness, existing works have not explored the application of 3D condition generation based on consistent flow matching to robotic manipulation. Therefore, we propose FlowPolicy, a conditional consistency flow matching model, which guarantees the generation of high-quality actions while also accomplishing one-step inference for real-time applications. This section is organized as follows: we first review flow matching and consistent flow matching, and then introduce the pipeline of FlowPolicy. Finally, we describe the design details of each component.

\subsection{Consistency Flow Matching}
We first introduce the notion of flow matching~\cite{lipman2022flow}, given a ground truth vector field $u(t, x)$ that generates the probabilistic path $p_t$ under two marginal constraints, $p_{t=0}=p_0$ and $p_{t=1}=p_1$. The expectation is to find an ODE whose solution transmits the noise ${x_0}\sim{p_0}$ to the data ${x_1}\sim{p_1}$:
\begin{equation} \label{eq:1}
     \left\{\begin{array}{rcl}
     \frac{d{\xi}_{x}(t)}{dt}={\nu}_{\theta}(t,{\xi}_x(t))\\{\xi}_x(0)=x\end{array}\right.
\end{equation}
${\xi}_x(t)$ is called a flow. The goal of flow matching is therefore to learn a vector field ${\nu}_{\theta}(t,x_t)$:
\begin{equation}\label{eq:2}
    L_{\text{FM}}(\theta)=E_{t,p_t}{\Vert{{\nu}_{\theta}(t,x_t)-u(t,x_t)}\Vert}^2_2
\end{equation}

Flow matching is a simple and attractive objective. However, lack of a prior knowledge about $u$ and $p_t$, conditional flow matching~\cite{lipman2022flow} is typically employed to regress ${\nu}_{\theta}(t,x_t)$ given a conditional vector field $u(t,x_t|{x_1})$ and a conditional probability path $p_t(x_t|{x_1})$:
\begin{equation}\label{eq:3}
    L_{\text{FM}^{\ast}}(\theta)=E_{t,q_{({x_1})}}E_{t,{(p_t|{x_1}})}{\Vert{{\nu}_{\theta}(t,x_t)-u(t,x_t|x_1)}\Vert}^2_2
\end{equation}

Where $x_1{\sim}q(x_1)$. Learning straight-line flows enables faster inference efficiency. Consistency flow matching (Consistency-FM) ~\cite{yang2024consistencyflowmatchingdefining} is a generalized method for efficiently learning straight-line flows without approximating the entire probabilistic path. Consistent velocity is defined as ${\nu}(t,{\xi}_x(t)) = {\nu}(0, x)$, indicating that the velocity along the solution of Equation \ref{eq:1} remains constant. Due to the complexity of the target distribution solution, Consistency-FM does not regress directly on the ground truth vector field, instead, it directly defines a velocity-consistent straight-line flow that departs at different times to the same endpoint. Consistency-FM can regress on the following objective to learn a velocity-consistent vector field:
\begin{multline}
    {L({\theta})=E_{t{\sim}\mu}E_{x_t,x_{t}+\Delta{t}}\|f_{\theta}(t,x_t)-f_{\theta^-}(t+\Delta{t},x_{t+\Delta{t}})\|^2_2}  \\
    {+\alpha{\|{\nu}_{\theta}(t,x_t)-{\nu}_{\theta^{-}}(t+\Delta{t},x_{t+\Delta{t}})\|^2_2}} \\
    f_{\theta}(t,x_t)=x_t+(1-t)*{\nu}_{\theta}(t,x_t)
\end{multline}

where $\mu$ is a uniform distribution on $[0,1-\Delta{t}]$, $\alpha$ is a positive scalar, and $\Delta{t}$ is the time interval, a small and positive scalar. ${\theta}^-$ denotes the running average of past $\theta$ values computed using an exponential moving average (EMA). Training $L(\theta)$ not only regularizes the velocity but also learns the data distribution.

Additionally, Consistency-FM allows for multi-segment training to enhance its ability to express the distribution of transmission through $p_t$. Specifically, multi-segment Consistency-FM splits time into uniform subsets, normalizes a consistent vector field ${\nu}^{i}_{\theta}{(t,x_t)}$ within each subset, and in turn constructs segmented linear trajectories. This approach relaxes the velocity consistency requirement and makes data transmission more flexible and efficient. Given the number of segments $K$, the training loss of segmented Consistency-FM is as follows:
\begin{multline}
    L({{\theta}^{\ast}})=E_{t{\sim}{\mu}^i}{\lambda}^iE_{x_t,x_{t}+\Delta{t}}\|f^i_{\theta}(t,x_t) \\ -f^i_{\theta^-}(t+\Delta{t},x_{t+\Delta{t}})\|^2_{2} \\
    {+\alpha{\|{\nu}^i_{\theta}(t,x_t)-{\nu}^i_{\theta^{-}}(t+\Delta{t},x_{t+\Delta{t}})\|^2_{2}}} \\
    f^i_{\theta}(t,x_t)=x_t+((i+1)/k-t)*{\nu}^i_{\theta}(t,x_t)
\end{multline}

Here, ${\mu}^i$ is a uniform distribution on $[i/K, (i + 1)/K-\Delta{t}]$. The moving step of each segment is controlled by a very small positive number $\Delta{t}$. $x$ and $x_t$ both follow a well-established distribution pattern. The flow and vector fields of each segment are described by ${\nu}^i_{\theta}(t,x_t)$, which depend on the parameters $\theta$ and time $t$. In addition, ${\lambda}^i$ is a positive weighting factor used to balance the importance of the different sections, especially since vector fields in the middle section are usually more difficult to train.

\subsection{FlowPolicy}
The overall pipeline of our proposed FlowPolicy is presented in Fig.~\ref{fig:pipeline}. FlowPolicy takes the powerful 3D diffusion policy (DP3) as the underlying architecture. DP3 is the state-of-the-art 3D vision-based conditional diffusion policy model. We address the inefficiency of 3D policy inference by constructing conditional consistency flow matching as a policy generator while ensuring the quality of task completion. Specifically, we consider the observed 3D point cloud as a visual conditional representation, which is combined with state embedding to guide the flow model to learn the manipulation policy. In the training phase, we regress the consistency vector field and learn the straight-line flow through velocity consistency loss, which directly generates the robot action sequences. Depending on the consistency training and condition representation, FlowPolicy can generate high-quality actions in one-step inference and accomplish complex manipulation tasks.

\begin{table*}[t]
    \centering
    \resizebox{1.0\linewidth}{!}{
    \begin{tabular}{
        l|
        >{\centering\arraybackslash}p{1cm}|
        >{\centering\arraybackslash}p{1.4cm}
        >{\centering\arraybackslash}p{1.4cm}
        >{\centering\arraybackslash}p{1.4cm}|
        >{\centering\arraybackslash}p{1.8cm}
        >{\centering\arraybackslash}p{1.8cm}
        >{\centering\arraybackslash}p{1.8cm}
        >{\centering\arraybackslash}p{2.2cm}|
        >{\centering\arraybackslash}p{1.8cm}}
    \toprule
    \multirow{2}{*}{Methods} & \multirow{2}{*}{NFE} & \multicolumn{3}{c|}{Adroit}  & \multicolumn{4}{c|}{Metaworld} & \multirow{2}{*}{\textbf{Average}} \\
     &  & Hammer & Door & Pen & Easy(21) & Medium(4) & Hard(4) & Very Hard(5) & \\  \midrule
    DP & 10 & 103.6$\pm${0.6} &102.5$\pm${0.4}&94.9$\pm${6.7}&104.9$\pm${2.2}&105.3$\pm${0.6}&120.3$\pm${1.4}&103.3$\pm${2.7}&106.1$\pm${2.4}\\
    DP3        & 10 &146.6$\pm${0.7}&146.8$\pm${2.0}&144.8$\pm${2.9}&145.9$\pm${2.1}&144.7$\pm${2.3}&148.6$\pm${3.4}&143.2$\pm${2.6}&145.7$\pm${2.3} \\
    Simple DP3 & 10 &64.5$\pm${1.4}&64.2$\pm${2.4}&71.2$\pm${9.3}&62.7$\pm${6.2}&59.4$\pm${4.1}&63.0$\pm${6.8}&65.3$\pm${6.7}&63.0$\pm${5.9} \\ \midrule
    \textbf{FlowPolicy (Ours)}             & 1  &20.4$\pm${0.5}&19.3$\pm${0.9}&20.6$\pm${0.2}&19.9$\pm${0.1}& 19.8$\pm${0.3}&20.1$\pm${0.2}&20.0$\pm${0.1}&\textbf{19.9}$\pm${\textbf{0.2}}\\ \bottomrule
    \end{tabular}
    }
    \caption{Quantitative comparison of runtime between state-of-the-art policy models~\cite{chi2023diffusion,ze20243ddiffusionpolicygeneralizable} and the proposed FlowPolicy. We evaluate 37 tasks from Adroit and Metaworld across 3 random seeds and report inference time per step (ms) with standard deviation. Our FlowPolicy enables real-time robot operations with an average time of 19.9ms in one-step inference, which is 7 $\times$ faster than DP3 and 3 $\times$ faster than its lightweight version, SimpleDP3.}
    \label{tab:results_runtime}
    \end{table*}

\begin{table*}[t]
    \centering
    \resizebox{1.0\linewidth}{!}{
    \begin{tabular}{
        l|
        >{\centering\arraybackslash}p{1cm}|
        >{\centering\arraybackslash}p{1.4cm}
        >{\centering\arraybackslash}p{1.4cm}
        >{\centering\arraybackslash}p{1.4cm}|
        >{\centering\arraybackslash}p{1.8cm}
        >{\centering\arraybackslash}p{1.8cm}
        >{\centering\arraybackslash}p{1.8cm}
        >{\centering\arraybackslash}p{2.2cm}|
        >{\centering\arraybackslash}p{1.8cm}}
    \toprule
    \multirow{2}{*}{Methods} & \multirow{2}{*}{NFE} & \multicolumn{3}{c|}{Adroit}  & \multicolumn{4}{c|}{Metaworld} & \multirow{2}{*}{\textbf{Average}} \\
     &  & Hammer & Door & Pen & Easy(21) & Medium(4) & Hard(4) & Very Hard(5) & \\  \midrule
    DP & 10 &16$\pm${10}&34$\pm${11}&13$\pm${2}& 50.7$\pm${6.1} & 11.0$\pm${2.5}& 5.25$\pm${2.5}& 22.0$\pm${5.0}& 35.2$\pm${5.3} \\
    Adaflow$\ast$ & - &45$\pm${11}& 27$\pm${6}& 18$\pm${6}& 49.4$\pm${6.8} & 12.0$\pm${5.0}& 5.75$\pm${4.0}& 24.0$\pm${4.8}& 35.6$\pm${6.1} \\
    CP & 1 &45$\pm${4}& 31$\pm${10} &13$\pm${6}& 69.3$\pm${4.2} & 21.2$\pm${6.0}& 17.5$\pm${3.9}& 30.0$\pm${4.9}& 50.1$\pm${4.7} \\
    DP3 & 10 &100$\pm${0}&56$\pm${5}&46$\pm${10}& 87.3$\pm${2.2} & 44.5$\pm${8.7}& 32.7$\pm${7.7}& 39.4$\pm${9.0}& 68.7$\pm${4.7} \\
    Simple DP3 & 10 &98$\pm${2} &40$\pm${17}&36$\pm${4}& 86.8$\pm${2.3}&42.0$\pm${6.5}&38.7$\pm${7.5}& 35.0$\pm${11.6}& 67.4$\pm${5.0} \\ \midrule
    \textbf{FlowPolicy (Ours)}             & 1  &100$\pm${0} &58$\pm${5}& 53$\pm${12}& 90.2$\pm${2.8}& 47.5$\pm${7.7}&37.2$\pm${7.2}& 36.6$\pm${7.6}& \textbf{70.0}$\pm${\textbf{4.7}} \\ \bottomrule
    \end{tabular}
    }
    \caption{Comparisons on success rate between state-of-the-art 2D-based~\cite{chi2023diffusion,hu2024adaflow,prasad2024consistencypolicyacceleratedvisuomotor} and 3D-based~\cite{ze20243ddiffusionpolicygeneralizable} policy models and the proposed FlowPolicy. We evaluate 37 tasks from Adroit and Metaworld across 3 random seeds and report the success rate (\%) with standard deviation. `$\ast$' indicates that the NFE of Adaflow is not fixed.}
    \label{tab:results_average_sr}
    \end{table*}

\subsubsection{3D visual conditional representation} FlowPolicy applies 3D point clouds as a guide to generate execution-specific policies given its rich spatial information. Specifically, we first acquire single-view depth-observation images and then convert the depth information into a point cloud via camera intrinsics and extrinsic. The single-view image is more practically applicable and the point cloud provides a richer spatial representation. Further, a farthest point sampling (FPS)~\cite{qi2017pointnetdeephierarchicalfeature} is applied to downsample the point cloud to preserve spatial wholeness and reduce redundancy. Finally, we employ a lightweight MLP to encode the point cloud into a compact 3D representation as the visual conditional representation for flow-based policy. With the conditional constraints of visual features, FlowPolicy enables the efficient capture of spatial information that is critical to fine robotic manipulation tasks.

\subsubsection{Conditional Trajectory Generation with FlowPolicy} For FlowPolicy, given the following definitions:
\begin{equation}
    a^{\text{src}} {\sim} \Gamma^{\text{src}}, a^{\text{tar}} {\sim} \Gamma^{\text{tar}}({\cdot}|s,v), t {\sim} \mu[\epsilon, 1-\epsilon]
\end{equation}

where $\Gamma^{\text{src}}$ is the source distribution, usually described as Gaussian noise $\Gamma^{\text{src}} = N(0,I)$. $\Gamma^{\text{tar}}$ is the target distribution conditioned by the state $s$ and visual representation $v$. $a^{\text{src}}$ and $a^{\text{tar}}$ are denoted as sequences of trajectories sampled from the source and target distributions. $t$ obeys a uniform distribution $t \in [\epsilon, 1-\epsilon]$. Define $a_t$ to be the interpolated trajectory between the source and destination trajectories at the transmission time step $t$. $F$ is the flow model conditioned by the state $s$ and visual representation $v$, FlowPolicy can be simply described as:
\begin{equation}
    \text{FlowPolicy}\ F: a,t,s,v\longrightarrow{\Delta{a}}
\end{equation}

To train a high-quality one-step generator, we employ a two-segment Consistency-FM and use backpropagation to minimize the following equation:
\begin{multline}
    L_{\text{FlowPolicy}}(\theta)=E_{t{\sim}{\mu}^i}{\lambda}^iE_{a_t,a_{t}+\Delta{t}}\|f^i_{\theta}(t,a_t,s,v) \\ -f^i_{\theta^-}(t+\Delta{t},a_{t+\Delta{t}},s,v)\|^2_2 \\
    {+\alpha{\|{\nu}^i_{\theta}(t,a_t,s,v)-{\nu}^i_{\theta^{-}}(t+\Delta{t},a_{t+\Delta{t}},s,v)\|^2_2}} \\
    f^i_{\theta}(t,a_t,s,v)=a_t+(i+1)/k-t*{\nu}^i_{\theta}(t,a_t,s,v)
\end{multline}

By training the two-segment loss above, two consistent vector fields can be learned, which in turn constructs a two-segment linear trajectory that transforms the noisy trajectory into the target trajectory. In the inference process, the noise trajectories are sampled from the source distribution $a_0 \in a^{\text{src}}$ and then predicted by a flow model to obtain action trajectories $a_1 \in a^{\text{tar}}$.

FlowPolicy provides a framework for generating high-quality actions in a one-step process that promotes a trade-off between efficiency and effectiveness.

\begin{figure*}[t]
\centering
\includegraphics[width=1.0\linewidth]{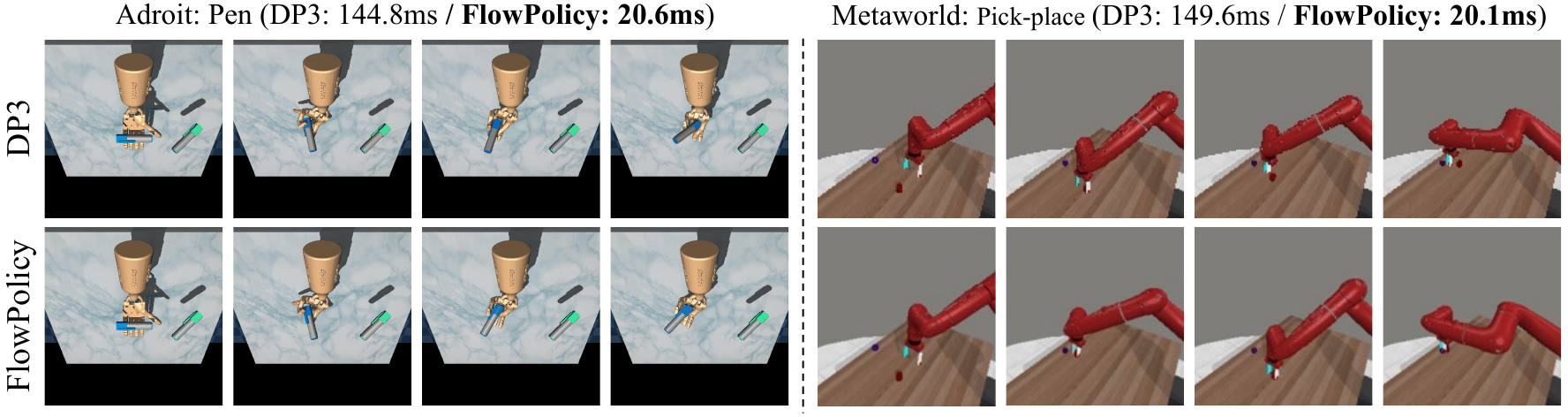}
\caption{Qualitative Comparison of FlowPolicy and DP3~\cite{ze20243ddiffusionpolicygeneralizable} on two challenging manipulation tasks from Adroit and Metaworld. Our method successfully generates high-quality actions at real-time speeds, completing these tasks effectively, whereas DP3 either produces lower-quality actions (left) or fails to complete the task (right).}
\label{fig:qualitative_results}
\end{figure*}

\section{Experiments}
\subsection{Dataset and Implementation Details}
\subsubsection{Simulation Benchmarks} We choose two preeminent environmental simulators, Adroit~\cite{rajeswaran2018learningcomplexdexterousmanipulation} and Metaworld~\cite{yu2021metaworldbenchmarkevaluationmultitask}, as our benchmarks for a comprehensive set of 37 tasks. The Adroit benchmark presents three challenges of dexterous manipulation tasks, where the execution of complex tasks is facilitated by the intricate control of a high-dimensional, articulated hand. In contrast, the Metaworld benchmark offers a diverse array of tasks that span the spectrum of difficulty levels, from easy to very hard, typically surmounted with the aid of a two-fingered robotic arm. For the procurement of expert demonstrations, we have harnessed a well-trained heuristic strategy, designed to yield datasets of exceptional quality. 

\subsubsection{Baselines} The main objective of this work is to leverage 3D point clouds as visual conditional representations, aiming to elevate the robot manipulation to the real-time level via a consistency flow matching strategy. Therefore, we focus on comparing DP3, a state-of-the-art 3D conditional diffusion-based strategy model. In addition, its lightweight version, simple DP3, is also compared. 
We also compared state-of-the-art 2D-based approaches, including diffusion policy (DP) \cite{chi2023diffusion}, Adaflow \cite{hu2024adaflow}, and Consistency Policy (CP)~\cite{prasad2024consistencypolicyacceleratedvisuomotor} and collected the same viewpoint images for them.
\subsubsection{Evaluation Metrics} We evaluate 20 episodes per 200 training epochs and record the mean of the five highest success rates. 
Each task is run repeatedly under three different random seeds, and their means and variances as well as inference times are calculated. In addition, the Number of Function Evaluations (NFE) is evaluated as a metric of inference efficiency. The higher the success rate, the better, and the lower the NFE and inference time, the better.
\subsubsection{Implementation Details} For each task, we generate a compact conditional embedding consisting of a 3D point cloud and the robot's state with a dimension of 64. Additionally, our model employs an observation horizon of two steps, signifying that it leverages the point clouds from the two most recent time frames as conditional inputs for policy generation. We assign ten expert demonstrations to each task. The image size is cropped to a resolution of 84$\times$84 pixels, and the point cloud is downsampled to 512 points for Adroit or 1024 points for Metaworld using FPS. During the training phase, the weights are updated using the AdamW optimizer with a learning rate of 1e-4 and a batch size of 128. The training process is iterated for 3,000 epochs. EMA decay rate is set to the default value of 0.95. Additionally, both action and state inputs are normalized to the interval [-1, 1] to facilitate the training process, with actions being unnormalized before execution. Our model and all compared methods are developed based on the PyTorch framework and evaluated on a single NVIDIA RTX 2080Ti GPU.

\subsection{Comparison with State-of-the-art Methods}
 
\subsubsection{Quantitative Comparisons on Runtime} Table~\ref{tab:results_runtime} reports the average inference time of each model for 3 dexterity tasks from Adroit and 34 different difficulty tasks from Metaworld. Our FlowPolicy demonstrates remarkable results by reducing the average inference time to 19.9ms, improving by more than 7$\times$ (19.9ms vs. 145.7ms) compared to the DP3 model. Even for the lightweight version of Simple DP3, FlowPolicy enhances the inference efficiency by over 3$\times$ (19.9ms vs. 63.0ms). The consistent velocity field provides a more direct linear path, enabling faster and more efficient action generation. Notably, our approach exhibits less variation in inference time. Specifically, FlowPolicy has less fluctuation in the average runtime for both individual and overall tasks, with a potential explanation attributed to the deterministic sampling policy of consistent flow matching. The results show that faster ODE trajectories can be determined and the inference efficiency can be improved by directly defining a straight-line flow, which is crucial for realizing real-time robot operation.

\begin{figure}[t]
\centering
\includegraphics[width=1.0\linewidth]{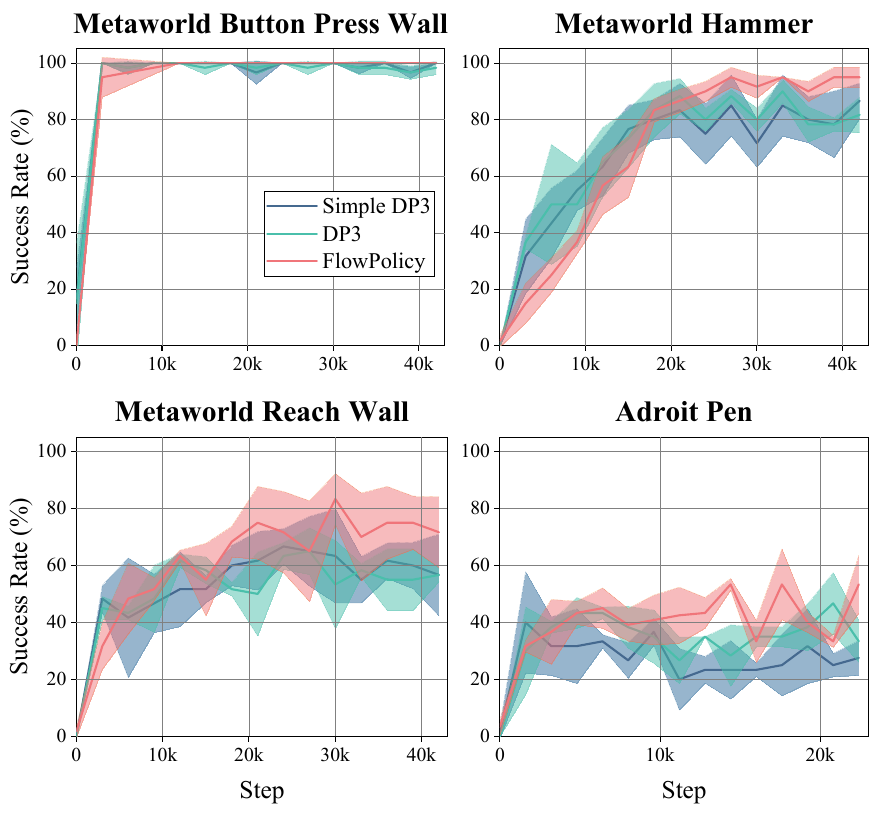}
\caption{Illustrations of the learning curves. Compared to Simple DP3 and DP3, FlowPolicy demonstrates higher stability, learning efficiency, and success rates.}
\label{fig:learning_curve}
\end{figure}

\subsubsection{Quantitative Comparisons on Success Rate} Further, Table~\ref{tab:results_average_sr} shows the average success rate of each model on 37 tasks in Adroit and Metaworld. 3D-based baselines generally outperform 2D-based ones, owing to the richer spatial representation offered by point clouds. Among 3D-based baselines, our approach achieves the best success rate with the shortest inference time. Specifically, FlowPolicy's overall success rate is 1.3$\%$ higher than DP3 and 2.6$\%$ higher than Simple DP3 with the same expert demonstration. The results show that FlowPolicy can achieve competitive results with state-of-the-art methods in just one step of inference. This is attributed to conditional trajectory generation based on consistent flow matching. On the one hand, the compact 3D representation leads to a richer spatial representation of the flow model. On the other hand, directly defining the straight-line flow motivates faster and simpler ODE trajectories to the extent that they can be generated in one step. Finally, multi-segment training ensures a trade-off between generation quality and efficiency. As a result, FlowPolicy decodes high-quality robot actions in just one step of inference, which demonstrates the effectiveness of consistency flow matching in robot manipulation tasks.

\begin{figure}[t]
\centering
\includegraphics[width=1.0\linewidth]{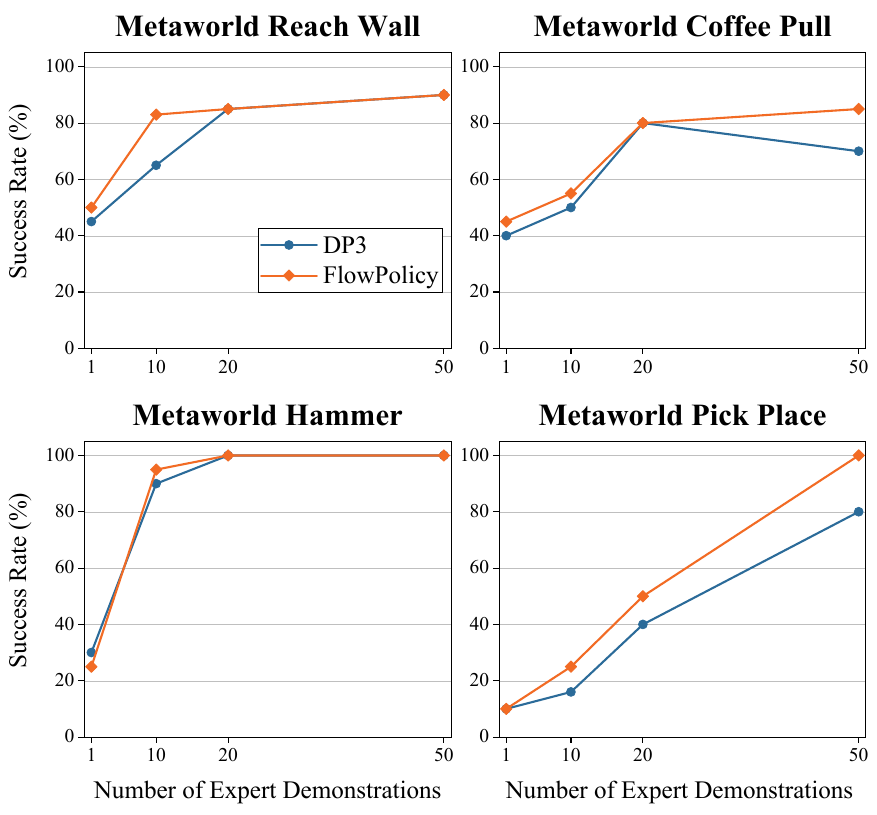}
\caption{Ablation on the number of expert demonstrations. We choose four typical tasks to explore the impact of different numbers of demonstrations on FlowPolicy and DP3. Both generally improve the accuracy with more demonstrations, but FlowPolicy typically has a higher success rate and avoids the performance bottleneck as presented in DP3.}
\label{fig:ablation}
\end{figure}

\subsubsection{Qualitative Comparisons on Manipulation Tasks} 
As illustrated in Figure~\ref{fig:qualitative_results}, we selected two typical qualitative examples in Adroit and Metaworld, generated by DP3 and FlowPolicy, respectively. From left to right, this sequence delineates the holistic progression of the task's execution, with each frame of the sequence corresponding to a distinct phase in the operation, as performed by the agent. On the left is a `Pen-turning' task involving complex dexterity manipulation, which requires turning the blue-capped pen in the dexterous hand to a specific position (i.e., the green-capped pen serves as the reference angle). Although DP3 accomplishes the dexterity task, the diffusion policy generated based on DP3 fails to ensure consistency with the target pen in 3D space compared to FlowPolicy. Specifically, DP3 turns one end of the pen to a state so high that it is darker in the camera view. In contrast, FlowPolicy generates higher-quality action sequences to turn the pen to the most appropriate angle. More importantly, FlowPolicy gets closer to the target position faster at the same frame. The right case is a difficult `Pick-Place' task in Metaworld, which requires the use of parallel gripping clips to carry a red cube to a blue target location. DP3 unsuccessfully picks up the red cube and fails the task. Instead, FlowPolicy correctly locates the position of the red cube and successfully carries it to the target location. The above findings show that our FlowPolicy not only accomplishes the tasks that DP3 cannot accomplish well, but also achieves a higher quality and speed of accomplishment than DP3. This further demonstrates that our proposed FlowPolicy has a more delicate understanding and generative ability in robot manipulation tasks.

\subsubsection{Analysis of Learning Curves} Similarly, we selected four representative tasks in Adroit and Metaworld and visualized their learning curves,  as illustrated in Figure~\ref{fig:learning_curve}. Although we defined 3000 epochs for each task, most of the tasks converged and leveled off at 500–1000 epochs. For the simple-level task `Button Press Wall', all three models learned the correct policy easily and converged with a small number of iterations. Although Simple DP3 and DP3 have high success rates, their success rates fluctuate as training progresses. On the contrary, our FlowPolicy learns the button policy well and maintains it, demonstrating the stability of FlowPolicy. For another simple level task 'Reach Wall' and a medium difficulty task `Hammer', all three models learn faster, but Simple DP3 and DP3 reach a performance bottleneck, while our method achieves a higher task success rate. The learning curves for the dexterity task in Adroit show similar conclusions. The above results show that FlowPolicy can learn policies faster and better, and demonstrates certain stability.

\subsection{Ablation Studies on Expert Demonstrations}
The success rate of the agent in accomplishing tasks depends on the number and quality of expert demonstrations, where the quality of demonstrations is ensured by a heuristic algorithm. We further conduct ablation studies to verify the influence of the number of expert demonstrations. As illustrated in Figure~\ref{fig:ablation}, the success rate variation curves for typical tasks with different number of demonstrations. For non-hard tasks (i.e., `Reach Wall', `Hammer'), FlowPolicy can outperform DP3 with a limited number of presentations. For hard-level tasks (i.e., `Pick-Place'), the success rate of the task can be significantly improved by increasing the number of expert presentations, as shown in both FlowPolicy and DP3. However, DP3 reaches performance saturation more easily, which even leads to a decrease in the success rate when the number of demonstrations is sufficient (i.e., `Coffee Pull'), while FlowPolicy performs better. The experiments reveal a better trade-off between the number of expert demonstrations and the success rate of the proposed FlowPolicy, i.e., the ability to learn effective data distributions from a small amount of data and to break through performance bottlenecks when the amount of data is sufficient.

\section{Conclusion}
In this work, we have introduced FlowPolicy, a novel 3D flow-based visual imitation learning algorithm. It can be summarized as a policy-generating model based on 3D point cloud and consistency flow matching. For the first time, we have explored the challenge of conditional generation within consistency flow matching, particularly under complex 3D visual representations. FlowPolicy accelerates the transfer of data from the noise to the action space by defining straight-line flows, thereby improving the inference efficiency of robotic manipulation tasks at the real-time level. Evaluations across 37 tasks of diverse challenges on Adroit and Metaworld platforms demonstrate that FlowPolicy achieves competitive success rates with real-time inference speed. We hope our work will advance 3D visual-based imitation learning towards practical applications.

\section{Acknowledgments}
This work was supported in part by National Natural Science Foundation of China (NSFC) under Grant Nos. 62372091, 62071097 and in part by Sichuan Science and Technology Program under Grant Nos. 2023NSFSC0462, 2023NSFSC0458, 2023NSFSC1972.

\bibliography{aaai25}

\end{document}